\documentclass[journal]{IEEEtran}
\ifCLASSINFOpdf
\else
\fi
\usepackage{amssymb}
\usepackage{amsmath}
\usepackage{amsfonts}
\usepackage{mathtools}
\DeclareMathOperator*{\argmax}{arg\,max}
\usepackage{xcolor}
\usepackage{hyperref}
\usepackage{multirow}
\usepackage{tabularx}
\usepackage{colortbl}
\newcommand \Tstrut {\rule{0pt}{2.6ex}}
\newcommand \Bstrut {\rule[-0.9ex]{0pt}{0pt}}

\begin{document}

\newcolumntype{C}[1]{>{\centering\arraybackslash}p{#1}}
%


\title{Pose Invariant Person Re-Identification using Robust Pose-transformation GAN}

%
%
%

\author{Arnab~Karmakar,\IEEEmembership{}
        and~Deepak~Mishra,~\IEEEmembership{Member~IEEE}
\thanks{A. Karmakar is with Human Spece Flight Centre, Indian Space Research Organisation, Bengaluru, Karnataka 560231, India,  e-mail: arnabkarmakar.001@gmail.com}
\thanks{D. Mishra is with Indian Institute of Space Science of Technology, Trivandrum, Kerala 695547, India, e-mail: deepak.mishra@iist.ac.in}
\thanks{Manuscript received March 19, 2020; revised April 26, 2021.}
}

\maketitle

\begin{abstract}
The objective of person re-identification (re-ID) is to retrieve a person's images from an image gallery, given a single instance of the person of interest. Despite several advancements, learning discriminative identity-sensitive and viewpoint invariant features for robust Person Re-identification is a major challenge owing to large pose variation of humans. This paper proposes a re-ID pipeline which utilizes the image generation capability of Generative Adversarial Networks combined with pose clustering and feature fusion to achieve pose invariant feature learning. The objective is to model a given person under different viewpoints and large pose changes and extract the most discriminative features from all the appearances. The pose transformational GAN (pt-GAN) module is trained to generate a person's image in any given pose. In order to identify the most significant poses for discriminative feature extraction, a Pose Clustering module is proposed. The given instance of the person is modelled in varying poses and these features are effectively combined through the Feature Fusion Network. The final re-ID model consisting of these 3 sub blocks, alleviates the pose dependence in person re-ID. Also, The proposed model is robust to occlusion, scale, rotation and illumination, providing a framework for viewpoint invariant feature learning. The proposed method outperforms the state-of-the-art GAN based models in 4 benchmark datasets. It also surpasses the state-of-the-art models that report higher re-ID accuracy in terms of improvement over baseline.

The code snippets\footnote{The  code  published  here  are  snippets  for  various  sub-parts  of  the  whole project.  Improvements  are  being  made  in  the  project  and  a  fully  executable code will be published soon.} of this work canbe found in \href{https://github.com/arnabk001/Pose-Invariant-Person-Re-Identification-using-Robust-Pose-Transfromation-GAN}{https://github.com/arnabk001/Pose-Invariant-Person-Re-Identification-using-Robust-Pose-Transfromation-GAN}

\end{abstract}

\begin{IEEEkeywords}
Person Re-Identification, Pose Transformation, Generative Adversarial Networks, Pose Invariance.
\end{IEEEkeywords}

%
\IEEEpeerreviewmaketitle

\section{Introduction}

\IEEEPARstart{I}{n} one of the earliest definitions~\cite{plantinga1961}, Person Re-Identification is described as, \emph{``to identify it (the person) as the same particular as one encountered on a previous occasion"}. In image based re-ID, the detected and localised set of pedestrians construct the ‘gallery’ set. Given any query image, the re-ID model aims to retrieve all possible instances of the person-of-interest from the gallery set. Therefore, person re-ID is considered as an interdisciplinary field which sits in between image classification and information retrieval.

\begin{figure}[htpb]
	\centering
	\includegraphics[width=0.98\linewidth]{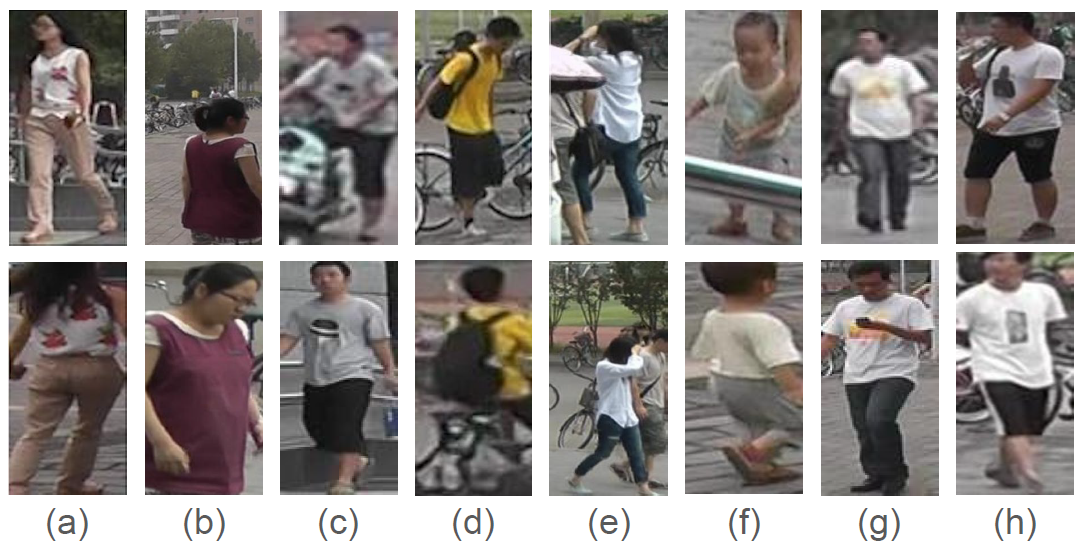}
	\caption{Challenges in ReID: (a-b) inaccurate detection, (c-d) pose misalignment, (e-f) occlusions, (g-h) very similar appearance of different IDs}
	\label{fig:challenges}
\end{figure}

In recent years, person re-ID has gained significant importance due to its widespread applications in security and surveillance, public safety, crowd control etc. Combining person re-ID with other branches of computer vision, such as face recognition, pose determination, activity recognition etc. head towards developing smart homes and cities. However, the primary bottleneck in developing robust person re-ID models arises from intra-class variations, i.e. the large appearance change of the same person across different scenes and camera views. This is mainly contributed to large pose change of the person-of-interest, which is also responsible for self-occlusion of the subject. The pose change of human subjects often lead to occlusion by external attributes (objects). In addition to that, in image-based re-ID, challenges such as inaccurate detection, occlusion, variation of illumination and resolution etc. are significant hindrances (Fig.~\ref{fig:challenges}) from achieving high accuracy with robustness.

Person re-identification is generally performed in two steps:
\begin{enumerate}
	\item learning discriminative features from the given image of the person-of-interest, and
	\item retrieve (and rank) all possible instances of the person-of-interest from the gallery set based on the feature similarity (distance metric)
\end{enumerate}

Most existing methods~\cite{ahmed2015improved}~\cite{deepreid} make use of complex Deep Neural Networks (DNNs) as the backbone for feature extraction from pedestrian images. However, DNNs require a large number of samples (images) per person for proper training, hence it has limited success on smaller datasets. Also, the conventional problems of varying illumination, scale and occlusion are still persistent with DNN based model.

The idea of pose independence has been recently explored in the re-ID community~\cite{johnson2018person}~\cite{posebox}~\cite{saquib2018pose}~\cite{bak2015person}. In this work, we argue that pose variation is the most significant hindrance when it comes to achieving highly accurate re-ID performance, and we focus on leveraging the pose dependence along with attaining invariance in occlusion, scale, illumination and rotation. We achieve this by modelling the person-of-interest into multiple significant poses and then combining the extracted features from each of these models to generate a complete viewpoint invariant feature vector. The proposed end-to-end re-ID pipeline is constructed upon three building blocks:
\begin{enumerate}
	\item A pose transformational GAN (pt-GAN) model to generate a given person image in any pose
	\item An unsupervised Pose Clustering model to select the significant poses for pose transformation
	\item A Feature Fusion Network (FusionNet) to effectively combine the discriminative features from the generated images and the source image.
\end{enumerate}

At the heart of this re-ID model is the in-house developed pose transformational GAN (pt-GAN), which is an improvement over our previous work~\cite{mywork} in terms of loss function and training methodology; finetuned on re-ID datasets. Combining this with Pose Clustering and FusionNet, we emphasize our contribution in this paper as follows:


\begin{enumerate}
	\item We propose an end-to-end feature extraction model for person re-ID while achieving pose invariance. Additionally, robustness is achieved in terms of illumination, scale, occlusion and rotation by using efficient data augmentation techniques and utilizing inherent CNN features through the GAN architecture.
	\item The proposed end-to-end trainable model provides a powerful methodology to learn a viewpoint invariant feature representation from a single instance of the person-of-interest.
	\item The proposed person re-ID framework incorporates an in-house developed pt-GAN architecture that serves as an efficient data augmentation tool. Incorporating this pose transformation GAN benefits in achieving an increase in the training dataset where the available dataset is small.
\end{enumerate}

Our work is validated on four benchmark datasets, i.e. three large scale public datasets Market-1501~\cite{market-1501}, CUHK03~\cite{cuhk03} and DukeMTMC-reID~\cite{dukemtmc-reid-usgan} and a small-scale dataset CUHK01~\cite{cuhk01}. \textbf{The proposed method surpasses the GAN based person re-ID models in both Rank-1 accuracy and mean Average Precision (mAP), and the results are comparable to the state-of-the art methods. Additionally, the proposed method also outperforms the popular state-of-the-art re-ID models in terms of performance improvement over baseline.}

The rest of the paper is organised as follows: in section~\ref{related-work}, recent practices and developments in the re-ID community is explained, drawing a comparison to our method. Subsequently in section~\ref{method}, we formally define the problem of person re-identification. The intrinsic details of our model is demonstrated in section~\ref{method} and the experimental setting and detailed results are compared in section~\ref{experiments} and \ref{results}, respectively.

\section{Related Work}\label{related-work}

Initial approaches of re-ID started with handcrafted features \cite{market-1501} \cite{farenzena2010person} \cite{weinrich2013appearance}, where a combination of manually extracted features are used as the image descriptor. Although these methods provide good accuracy in smaller datasets, there have not been any significant improvement of these methods over the years. Another direction of work focus on developing improved distance metric learning methods~\cite{bak2015person} to construct discriminative subspace for the extracted features for better matching accuracy. However, the scope of distance metric in improving re-ID performance is limited when the dataset is large, unless a powerful feature extractor is employed. A comprehensive overview of the history and development of re-ID methodologies is described by Zheng et al.~\cite{reid-ppf}.

With the release of larger datasets and higher computational capability, Deep Learning based methods have provided significant performance improvement in recent years ~\cite{reid-ppf}~\cite{sun2018beyond}. A
large portion of the existing methods depend heavily on the CNN classification backbone such as ResNet~\cite{resnet}, DenseNet etc. In this transfer learning setting, the backbone CNN is finetuned using the re-ID data using the traditional classification loss~\cite{ahmed2015improved}, or using triplets in a Siamese learning setting~\cite{cheng2016person}~\cite{varior2016gated}. Some of the methods experiment with a part-based feature extraction and matching in addition with full image features~\cite{cheng2016person}. Any of these methods, however, do not solve the problems of occlusion, background clutter, inaccurate detection, varying scale etc. Since deep learning methods give a holistic feature including the background, many methods use foreground segmentation or body-part based semantic segmentation to improve performance~\cite{posebox}, which leads to computational overhead. Many attempts of combining the handcrafted features with the CNN based features have also been observed~\cite{johnson2018person}~\cite{lee2016ensemble} since deep learning based method are not very successful in smaller datasets.

\subsection{Pose based methods in re-ID}
Exploring the human body pose to improve re-ID performance has been a recent development. Many previous works have analysed the dependence of human pose in person re-ID~\cite{johnson2018person}~\cite{posebox}~\cite{saquib2018pose}~\cite{bak2015person} In earlier works, the Symmetry-Driven Accumulation of Local Features (SDALF) method was proposed by Farenzena et al.~\cite{farenzena2010person} where handcrafted complementary features are extracted depending on localization of perceptually relevant human parts. They achieve pose invariance by combining these features through a weighting scheme based on human body (vertical) axis of symmetry. Weinrich et al.~\cite{weinrich2013appearance} detect the upper body pose for tracking the human subject in egocentric image frames. The local texture features of the upper body are learned and a generative 3D shape model is used for re-identification purpose. In another direction of research, there have been attempts to apply metric learning, based on a transformation function for different camera pairs, that can be used to model the pose change of the human and then compute feature distances to boost person re-ID. Following this philosophy, Bak et al.~\cite{bak2015person} try to learn a metric pool, i.e. a pose change metric based on Mahalanobis distance by classifying human pose into 3 groups: front, back and side. All the aforementioned methods use typical handcrafted features to achieve pose variation. However, in recent times there have been increasing application of deep learning based feature extractors and pose detectors for re-ID. Zheng et al.~\cite{bak2015person} addresses the pose misalignment problem by ‘PoseBox’, a pose based body part extraction and reconstruction algorithm for the pedestrian images. The original image and the PoseBox output both are trained in a CNN to extract pose invariant features for re-ID. Cho et al.~\cite{cho2016improving} proposes a multi-shot re-ID framework by learning the transformation matrix from one pose to another by utilizing external camera parameters to predict the next pose (target pose) of the pedestrian and then use the matching score for re-ID framework. Sarfraz et al.~\cite{saquib2018pose} uses the detected pose keypoints with the pedestrian image as the combined input to the CNN for learning pose sensitive embedding. The authors use a view predictor, i.e. pose classification branch in conjuncture with the CNN branch for robust embedding generation. However, the improvement of the framework is largely contributed to the new re-ranking methodology. Zhao et al.~\cite{spindlenet} proposes the ‘SpindleNet’ architecture that uses a Region Proposal Network (RPN) utilizing the human body structure (pose), and then extract features from each body part using the ROI pooling. These part-based features are merged using a tree-like ‘competetive’ feature fusion network with the full-image feature. Following this idea, Jhonson et al.~\cite{johnson2018person} tackles the person detection error by a SpindleNet-like body part based deep feature learning methodology by using an ensemble of handcrafted features such as Hue-Saturation-Value (HSV), Scale Invariant Local Ternary Pattern (SILTP) with the deeply learned model to produce better results. The fusion of Deep learning based features with handcrafted features have also been observed in Lee et al.’s~\cite{lee2016ensemble} work, where the authors use an ensemble of invariant features for re-ID, by combining holistic (deep) features and regional (handcrafted) features. However, the authors do not provide specific justification on how these features are pose invariant.

\subsection{GANs in Person re-ID}
Goodfellow et al.~\cite{goodfellowgan} introduced GANs for image generation from random noise. Recently, the idea of utilizing GANs for improving person re-ID has been explored by the re-ID community. In the initial methods, GANs were used as a data augmentation technique to provide sufficient training data to strengthen the learning process. Zheng et al.~\cite{dukemtmc-reid-usgan} achieve an improvement over baseline using the GAN generated data and label smoothing regularization to train the CNN model in a semi-supervised setting. Liu et al.~\cite{liu2018pose} proposes a pose transferrable GAN using a Generator-Guider-Discriminator architecture. A U-net~\cite{unet} based structure was adopted for generator and the vanilla CNN structure was followed for the guider-discriminator model. The authors randomly sample poses from the MARS dataset and generate samples with new poses to increase the dataset size to finetune re-ID performance using the backbone CNN network.

Two significant works that incorporate pose invariance using GANs in person re-ID, are Qian et al.~\cite{pngan} and Ge et al.~\cite{fdgan}. Ge et al.~\cite{fdgan} propose a Siamese learning structure using identity discriminator and pose discriminator as well as verification loss. This helps in learning identity related and pose unrelated feature embedding which they utilise through the backbone (ResNet-50 ~\cite{resnet}) network during feature generation while testing. Qian et al.~\cite{pngan} train a pose transformation GAN and for every query image, generate 8 canonical pose transformed images. It uses ResNet~\cite{resnet} architecture to extract image features from the source image and generated images. 
However, the authors use 8 canonical poses for every image and combine the feature vectors by element wise maximum operation. Both these decisions are fairly crude and the authors do not provide enough justification for using this type of a learning setting. In our work, we explore the effect of the number of generated images (with new poses), and apply a fully connected neural network FusionNet for effective combination of features. Our results on several benchmark datasets validate this claim and we achieve the best results among all GAN based models for re-ID.

\section{Methodology}\label{method}

\subsection{Definition}
Considering a closed world model, where a total of $n$ images define the gallery set $\mathbb{G} : \{ g_j \}_{j=1}^{n}$ having $\mathcal{N}$ different identities. Given a query (probe) image $q_i$ with identity $i$, the person reidentification framework aims to retrieve all possible instances of the same identity $i$ from the given gallery set $\mathbb{G}$
\begin{equation}
\mathcal{R}_i^\ast = \argmax_{i \in 1, 2, \dots, \mathcal{N}}~sim (q_i, \{g_j\}_{j=1}^n)
\end{equation}
where $\mathcal{R}_i^\ast$ is the retrieved set of all images with the same identity $i$ denoted by the probe image $q_i$ and $sim(\cdot,\cdot)$ denotes some form of similarity function.

Assuming a training dataset of $N$ images, denoted by $\mathcal{D}_{tr} = \{ I_i, y_i \}_{i=1}^N$, where $I_i$ denotes the person image and $y_i$ denotes its identity, the idea is to learn a feature extraction function $\Phi$ in order to represent a given image $I_i$ as a feature vector $\mathbf{f}_{I_i}=\Phi(I_i)$.

During testing, we need to retrieve all possible images of the same identity as the given probe image $I_q$ from the testing (gallery) set $\mathcal{D}_{te}=\{ I_j \}_{j=1}^{N'}$ based on the feature similarity measure $sim(\mathbf{f}_{I_q},\{\mathbf{f}_{I_j}\}_{j=1}^{N'})$.

\subsection{Framework Overview}
During training, the input image $I_i$ is transformed using ResNet50-1 model $\mathbf{F}_{R1}(\cdot)$ for the feature vector representation $\mathbf{F}_{R1}(I_i)$. This ResNet50-1 model is essentially a ResNet50 model trained on the ImageNet dataset.

The pose of $I_i$ is detected by an off-the-shelf pose detection model Openpose~\cite{openpose} and the pose vector is denoted as $p_i$. The pose clustering module analyses all detected poses in the dataset to generate a set of $K$ significant body poses 
$$\mathcal{P}_K = \{{p'}_{k} \}_{k=1}^K =\{ {p'}_{1}, {p'}_{2}, \dots , {p'}_{K}\} $$ 
The term ${p'}_{k}$ denotes the $k^{th}$ generated pose. The generator takes the input image feature $\mathbf{F}_{R1} (I_i)$ and the set of poses $\mathcal{P}_K$ to generate $K$ images,
\begin{align*}	
	\hat{I}_{i, \mathcal{P}_K} &= \mathbf{G}\big(\mathbf{F}_{R1} (I_i), \mathcal{P}_K\big)\\
	\implies \hat{I}_{i,{p'}_{k}}|_{k=1}^K &= \mathbf{G}\big(\mathbf{F}_{R1} (I_i), {p'}_{k}\big)|_{k=1}^K
\end{align*}
$\hat{I}_{i, \mathcal{P}_K}$ denotes the original image $I_i$ is being transformed to $K$ poses. Now, the ResNet50-2 model $\mathbf{F}_{R2}(\cdot)$ is used to transform the $K$ generated images into feature vectors $\mathbf{F}_{R2}(\hat{I}_{i, \mathcal{P}_K})$. 

The ResNet50-1 model is pre-trained on the ImageNet dataset, whereas the ResNet50-2 model is finetuned on the re-ID dataset. Initially while training the pt-GAN model, the general image features of the person is of primary concern, and preserving the image features as a whole would give a better reconstruction. Hence a vanilla ResNet50-1 is deployed to extract the general features of the person. In the second stage, since the features of the original image along with generated images is more important for re-ID matching, we have used the ResNet50-2 model which is finetuned for re-ID purpose.
 
The $K$ image feature vectors along with the input image feature $\mathbf{F}_{R1}(I_i)$ is transformed using the FusionNet model $\mathbf{F}_{FN}(\cdot)$ as the final feature vector. The complete transformation model, in accordance with the previously described notation, becomes

\begin{equation}
\mathbf{f}_{I_i} = \Phi(I_i) = \mathbf{F}_{FN}( \mathbf{F}_{R1}(I_i), \mathbf{F}_{R2} \big(\hat{I}_{i, \mathcal{P}_K})\big)
\end{equation}

\begin{figure*}[!ht]
	\centering
	\includegraphics[width=0.95\linewidth, ,height=6.5cm]{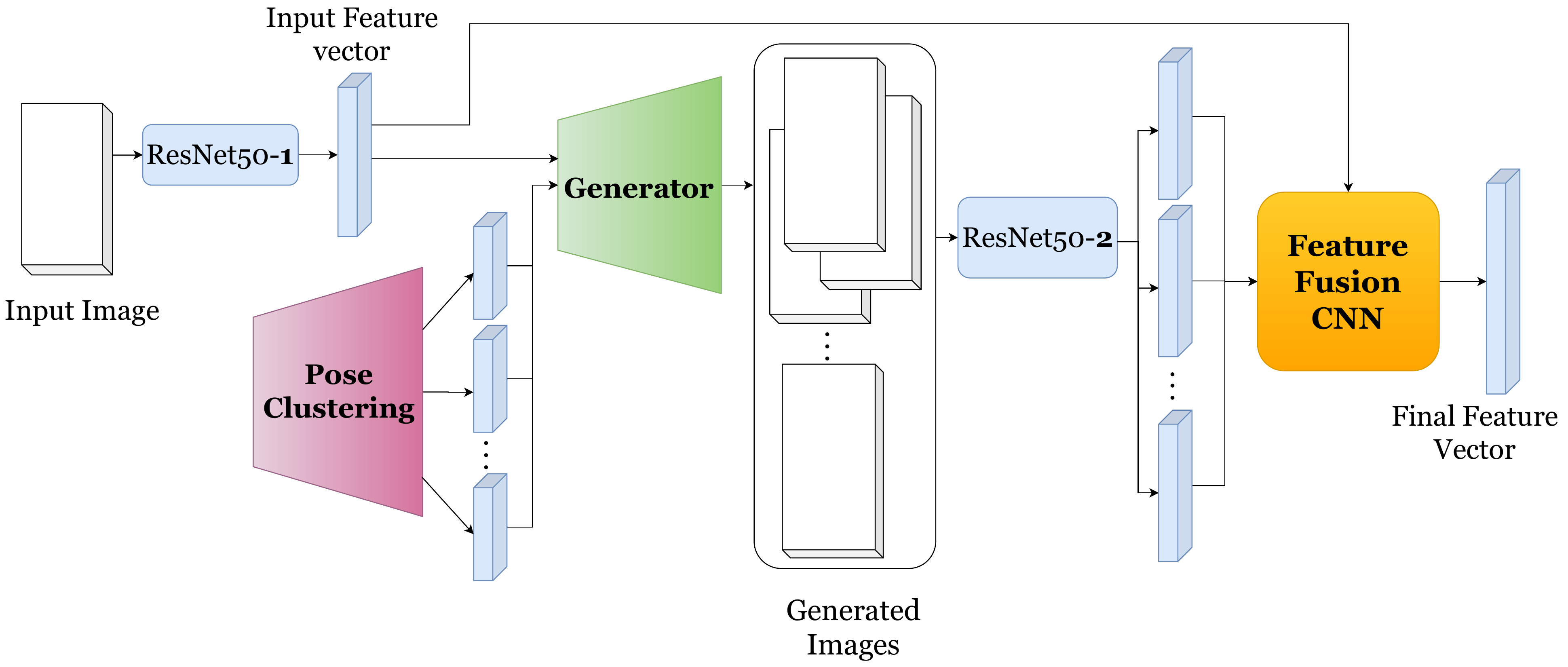}
	\caption{Full architecture of the proposed person reID pipeline, consisting of the trained generator from our pt-GAN model, pose clustering and FusionNet. The number of generated images vary according to the pose clustering algorithm. The FusionNet is trained keeping the Generator and the Pose Clustering module fixed. During testing the features are extracted from second last layer of FusionNet.}
	\label{fig:fullarch}
\end{figure*}

This feature vector $\mathbf{f}_{I_i}$ extracted from all images in the gallery set, construct the feature space, where image retrieval is performed based on a similarity function $sim(\cdot,\cdot)$. The complete architecture is depicted in Figure~\ref{fig:fullarch}.

\subsection{Pose Guided Person Image Synthesis}

In general, Generative Adversarial Networks (GANs) comprise of two parts: the Generator and the Discriminator. The Generator learns to approximate the underlying real data distribution to a random distribution, where sampling is performed to generate new samples. The Discriminator learns to distinguish the real life data and the generated data, thereby providing a measure of ‘goodness’ on how well the generator can mimic the real data. They both learn in a competitive manner.

In the pt-GAN model, we want the generator to learn the pedestrian image distribution, conditioned on human body pose. Our pt-GAN model is given an image pair $\{I_i, I_j\}$ of the same identity but different pose. The OpenPose~\cite{openpose} pose detection framework is used off-the-shelf to generate the pose vector $p_j$ of the target image $I_j$. The generator takes in the image feature vector $\mathbf{F}_{R1}(I_i)$, concatenated with the target pose vector $p_j$ to model the person in the desired pose
$\hat{I}_{j} = \mathbf{G}(\mathbf{F}_{R1}(I_i),p_j) $

The generator learns the data distribution to minimize the difference between the generated image $\hat{I}_{j}$ and the target image $I_j$. The loss function of the GAN model is denoted by,
\begin{align}
\mathcal{L}(\mathbf{D}, \mathbf{G}) &= \mathbb{E}_{I_j\sim p_{data}(I_j)}[\log(\mathbf{D}(I_j))] \nonumber \\ & \qquad +\log(1-\mathbf{D}(\mathbf{G}(\mathbf{F}_{R1}(I_i),p_j)))
\end{align}

We also use the added L2 norm for better reconsruction and cleaner image generation to the generator loss function.
\begin{equation}
	\mathcal{L}_{L_2} = \mathbb{E}_{I_j\sim p_{data}(I_j)}|| I_j - \hat{I}_j ||_2
\end{equation}

For Discriminator, we have added an extra classification branch in order to predict the class labels in addition to the real/fake classification. Incorporating classification loss in the discriminator complements the generator’s capability to reconstruct detailed images with lesser artifacts. The categorical crossentropy loss is used here,
\begin{equation}
	\mathcal{L}_c = - \sum_{i=1}^{n} \sum_{c=1}^{C} y_c \log(y'_c)
\end{equation}
where $y'_c$ is the predicted probability of class $c$, $y_c$ is the actual class label; summed over all classes across all samples.
Therefore we write the final loss function of our pt-GAN model as
\begin{align}
\mathcal{L}(\mathbf{D}, \mathbf{G}) =& ~\mathbb{E}_{I_j\sim p_{data}(I_j)}[\log(\mathbf{D}(I_j))] + \mathcal{L}_c\nonumber\\ & \quad + \log(1-\mathbf{D}(\mathbf{G}(\mathbf{F}_{R1}(I_i),p_j))) + \mathcal{L}_{L_2}
\end{align}

\begin{figure}[!ht]
	\centering
	\includegraphics[width=0.99\linewidth]{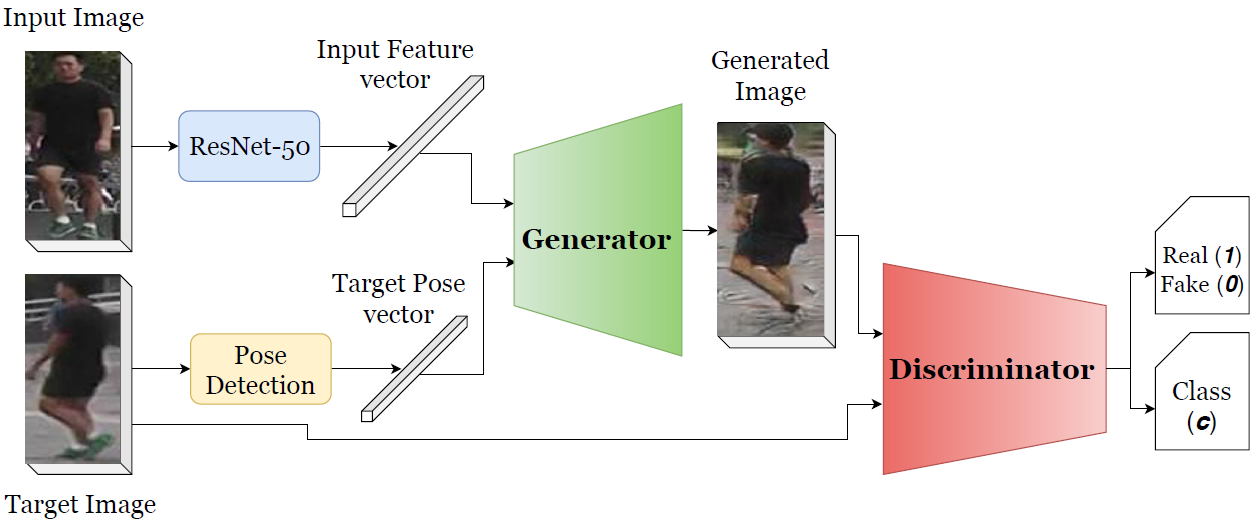}
	\caption{The complete pt-GAN model architecture. The Generator learns pedestrian features from the image descriptors extracted by the ResNet50-1 model and reconstructs the pedestrian into the target pose vector. An extra classification branch is added to the Discriminator that complements the Generator's capability to reconstruct detailed images with lesser artifacts.}
	\label{fig:ganfull}
\end{figure}

The GAN model is depicted in Figure~\ref{fig:ganfull}. This is an improvement over our previous work \cite{mywork} in terms of loss function formulation and training methodology, tuned for better performance in re-ID datasets. This model have been proven to produce robust human images with significant details. The image generation results are shown in section~\ref{results-image-gen}.

\subsection{Pose Clustering}
After our pt-GAN model is trained to generate pedestrian images in any given pose, the requirement is to select the minimum number of best possible poses to synthesize the least number of images but still be able to extract viewpoint invariant features. Clustering techniques are applied to study the maximally occurring poses in the dataset, as they provide an unsupervised approach to structure the data points by extracting meaningful insights about the distribution. In this work, we have implemented clustering on the pose vectors based on two criteria, (1) Full-Body Pose based clustering, (2) Body joint based clustering method. For each of these methods, we implement two major clustering algorithms: (i) K-means Clustering and (ii) Gaussian Mixture Model based Clustering (GMM Clustering). The K-means method produces discretized output points while GMM produces a distribution of data points, where sampling is performed to produce output poses.
\subsubsection{Full-body pose based Clustering}\label{fb_cl}
The complete pose vector $p_i$ is taken as a single data point and then clustered into $n_{cp}$ clusters. The K-means method directly produces the desired number of poses as the cluster centers. For image generation, these cluster centers are taken as the sample pose. Meanwhile the GMM method provides a pose distribution, and the required number of poses are randomly sampled and fed to the subsequent pt-GAN model to synthesize the pedestrian images.

\subsubsection{Body joint based Clustering}\label{bj_cl}
In this approach, each body joint location is taken individually, and clustering is performed in every body joint position with $n_{cbj}$ clusters, for $j^{th}$ body joint position. 
For K-means, the cluster centers are selected randomly out of $n_{cbj}$ clusters, for each body-joint, and the output pose is constructed as the accumulation of these body-joint co-ordinates.
For GMM, full body pose is constructed using individual body-joint co-ordinate that are randomly sampled from the distribution of individual body-joints.

\subsection{Feature Fusion Network}
The Pose Clustering module gives the $N_{aug}$ best possible poses for the given image, and the pt-GAN module transforms the input image into these poses. Having these many samples augmented from one image, we need an efficient way of combining the best possible features of every image. We propose a Feature Fusion Network (FusionNet) to combine the features of the input image with all the augmented image.

\begin{figure}[!ht]
	\centering
	\includegraphics[width=0.8\linewidth]{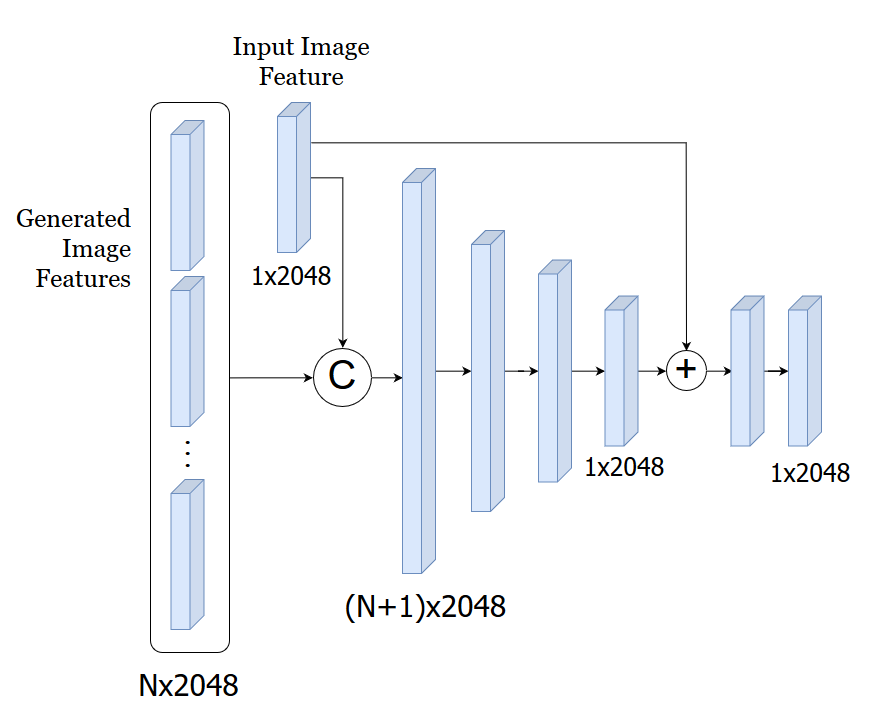}
	\caption{Architecture of the proposed FusionNet. $N$ stands for the number of generated images (correspondingly image feature vectors), $N \in \{ 8,12,16,24 \}$. The last layer of FusionNet is modified to the number of classes during training, while the features are extracted from second last layer during testing.}
	\label{fig:fusionnet}
\end{figure}

As depicted in the architecture of the FusionNet in Figure~\ref{fig:fusionnet}, the image descriptors of the input image as well as the generated images are obtained using a ResNet50-2 model trained (finetuned) on re-ID dataset. We first perform concatenation of all the image descriptors to obtain an $(N_{aug} + 1)$ length input feature vector which is the input to the fully connected FusionNet. Since the input image holds the most prominent features for re-identification and the generated images might incorporate various distortions, we introduce a skip connection with the input feature to re-use the input features and imply more importance to the original image, as shown in Figure~\ref{fig:fusionnet}. During training, the final layer of this network is modified to predict the class labels and this model is trained on classification loss (categorical crossentropy).

\section{Experiments}\label{experiments}

\subsection{Datasets}
\subsubsection{Market-1501~\cite{market-1501}}
This dataset contains 32,668 images of 1,501 identities, captured using 6 different cameras. We follow the standard split~\cite{market-1501}, where 12,936 images of 751 identities are used for training and the rest 19,732 images of 750 identities are used for testing. This dataset (training set) is used to train our pt-GAN model and the FusionNet model.
\subsubsection{DukeMTMC-reid~\cite{dukemtmc-reid-usgan}}
This dataset contains 36,411 images of 1,812 identities. We use 702 identities as the training set and the remaining 1,110 identities as the testing set, following the standard evaluation protocol~\cite{dukemtmc-reid-usgan}. 
\subsubsection{CUHK03~\cite{cuhk03}}
This dataset is captured using 6 cameras, containing 14,096 images of 1,467 identities. For training, 1367 identities are used. Validation and testing set consists of 100 identities each. We follow~\cite{cuhk03} for the testing process with 20 random splits.
\subsubsection{CUHK01~\cite{cuhk01}}
This is a small scale dataset compared to the other 3, having 971 identities with 2 images per person. As in~\cite{cuhk01}, the images of camera A is used as the probe image and camera B images are used as the gallery set.

\subsection{Evaluation Metrics}
The Cumulative Matching Curve (CMC) is generally used for evaluating the re-ID accuracy. However, for most of the existing methods the full details of the CMC curve is not shown, rather the rank-1, rank-5 and rank-10 accuracies are reported. Zheng et al.~\cite{market-1501} proposes to use the mean Average Precision (mAP) as a single measure for overall correctness of re-ID models. In this work, we report the r-1, r-5 and r-10 accuracies for all 4 datasets, and the mAP for Market-1501~\cite{market-1501}, DukeMTMC-reID~\cite{dukemtmc-reid-usgan} and CUHK03~\cite{cuhk03}.

\subsection{Implementation details}
\subsubsection{Data Augmentation}
Following~\cite{mywork}, the image augmentation techniques are used as is. Specifically, Random Erasing~\cite{randomerasing} is used for occlusion invariance, and Random Crop is used as an augmentation for inaccurate detections. Also, random rotation and crop in the range [$+20~deg , -20~deg$] is applied during the training of pt-GAN. Other methods such as random color jitter in all 3 channels, horizontal flip and random distortion are also incorporated while training the GAN model.
\subsubsection{GAN training}
For stable training and improving the performance of GANs, label smoothing is used with uniform random noise for the discriminator. The LeakyRelu activation and Adam optimizer (with $\beta_1=0.5$ and $\beta_2=0.999$) is used for both Generator and Discriminator. We have used the Kaiming-Normal initialization~\cite{resnet} for the GAN model and FusionNet. The other parameters such as the number of Residual blocks, the learning rate and the batch size is kept the same as our previous work~\cite{mywork}.\par
The re-ID images were resized to $128\times64$ for training. The number of poses $\mathcal{P}_K$ generated using the pose clustering module is varied from 8 to 24, i.e. $K\in \{8,12,16,24\}$, and the ablation studies are presented in Table~\ref{table:ablation}. For full body pose based clustering, the number of clusters is equal to the number of generated poses ($n_{cp}=K$), whereas for body joint based clustering, each body joint is clustered into 3 clusters ($n_{cbj}=3,\text{ for each bodyjoint }  j\in\{0,1,2,\cdots, 24\}$). For K-means, one out of 3 coordinates are randomly selected while generating full body pose, where for GMM, it is sampled from the distribution.\par
\subsubsection{Model Parameters}
The feature size is taken as $2048$ (output of the global average pooling layer of ResNet50) and the same is also replicated in the FusionNet architecture where the concatenated features are taken as input. The details of the FusionNet architecture is provided in Table~\ref{tab:fusionnet_arch}.

\begin{table}[htpb]
	\caption{Details of FusionNet architecture. For the optimal case GMM-12, $N=12$.}
	\label{tab:fusionnet_arch}
	\begin{tabular}{C{1.4cm} | C{1.4cm} C{1.4cm} C{1.4cm} C{1.1cm}}
		\hline
		\textbf{Layer Name}   & \textbf{Input feature size} & \textbf{Connected to}    & \textbf{Output feature size} & \textbf{Params (M)} \\
		\hline
		input\_img \arrayrulecolor{lightgray}  & -                  & ResNet50-1               & 2048                & -          \\ \arrayrulecolor{lightgray}\hline
		gen\_img\_ft & -                  & ResNet50-2               & N*2048              & -          \\ \hline
		concat\_1    & 2048, N*2048       & input\_img, gen\_img\_ft & (N+1)*2048          & -          \\ \hline
		fc\_1        & (N+1)*2048         & concat\_1                & 4*2048              & (N+1)*16.7 \\ \hline
		fc\_2        & 4*2048             & fc\_1                    & 2048                & 16.7       \\ \hline
		add\_1       & 2048               & fc\_2, input\_img        & 2048                & -          \\ \hline
		output       & 2048               & add\_1                   & 2048                & 4.1       \\
		\arrayrulecolor{black}\hline
	\end{tabular}
\end{table}

The FusionNet model contains a large number of trainable parameters. In order to prohibit overfitting, we have used dropout$=0.6$ along with Weight Regularization, Batch Normalization and Early Stopping. Both the pt-GAN model as well as the pose clustering module is trained independently. After training, the FusionNet model is trained for classification accuracy keeping the other two models frozen.


\section{Results and Discussion}\label{results}

\subsection{GAN Image Generation}\label{results-image-gen}

\begin{figure}[htpb]
	\centering
	\includegraphics[width=0.98\linewidth]{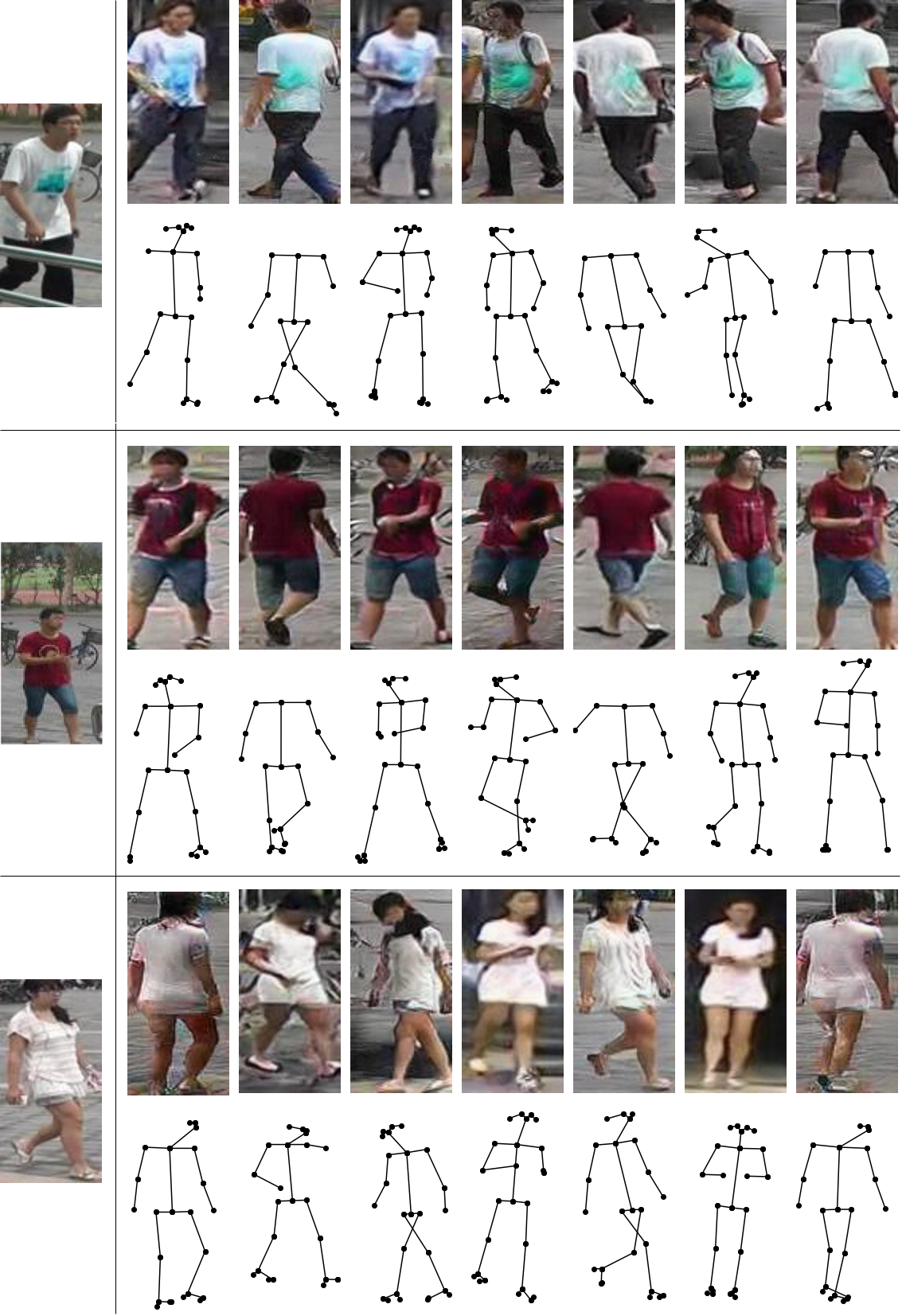}
	\caption{Generated Image of our pt-GAN model. The original image is shown in the left. The generated images along with the given target pose is shown subsequently.}
	\label{fig:gandemo}
\end{figure}

The proposed GAN model is trained on re-ID datasets to generate a pedestrian image into any given pose, given only a single instance of the person-of-interest. Figure~\ref{fig:gandemo} demonstrates the generated images from our pt-GAN model. The original image of the pedestrian is shown in the left, and the generated images with the target poses are shown in the right. As it can be seen, our pt-GAN model is able to correctly identify and extract the significant features from the input image and then reconstruct the pedestrian appearance in the target pose. Features such as clothing, hair, male/female attributes are properly retained.

\subsection{Pose Clustering}

\begin{figure}[htpb]
	\centering
	\includegraphics[width=0.98\linewidth]{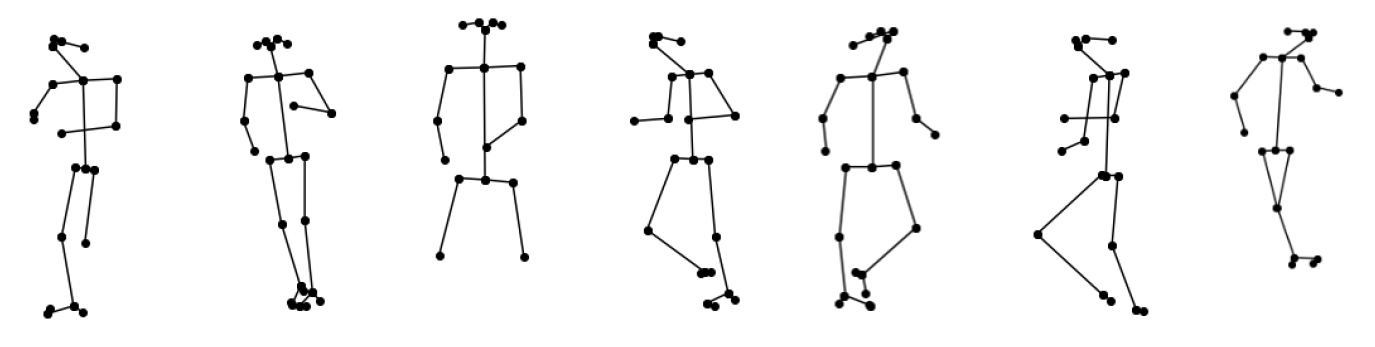}
	\caption{Sample Poses obtained using K-means Clustering with full-body pose based method.}
	\label{fig:kmeans_demo}
\end{figure}

\begin{figure}[htpb]
	\centering
	\includegraphics[width=0.98\linewidth]{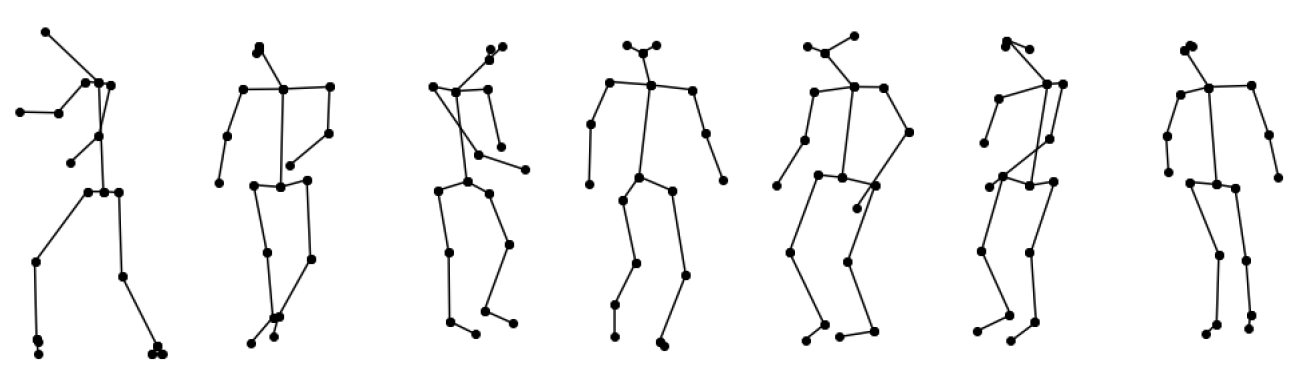}
	\caption{Sampled Poses obtained using GMM with full-body pose based	method}
	\label{fig:gmm_demo}
\end{figure}

The pose clustering model provides a variety of poses to augment the input image. Figure~\ref{fig:kmeans_demo} shows the poses obtained using K-means clustering in the Market-1501 dataset. Although K-means produces a fixed number of poses (according to the specified number of cluster centers), we achieve randomness when GMM is applied. Figure~\ref{fig:gmm_demo} shows a sample of generated images using GMM. In the final model, we have used GMM as it is able to provide superior results with less number of poses.


\subsection{Re-ID performance}
\subsubsection{Ablation studies}
We compare the results of both pose based and body-joint based clustering algorithms with number of poses varying from $8$ to $24$. As seen in Table~\ref{table:ablation}, K-means provides consistent performance in both the methods. The performance of GMM drops in body-joint based clustering algorithm as the distribution of each body joint is significantly sparse, when computed across all samples. We have observed the highest rank-1 accuracy in K-means-16, but the GMM-12 has been selected as the final model as it achieves higher mAP and comparable rank-1 accuracy with less number of poses.

\begin{table}[htpb]
	\centering
	\caption{Ablation study of the proposed model on Market-1501 dataset. The GMM-12 has been selected as the final model as it achieves higher mAP and very similar rank-1 accuracy with less number of poses compared to K-means-16.}
	\label{table:ablation}	
	\begin{tabular}{c|c|c ||c l}
		\hline
		\multicolumn{2}{c}{\parbox{2.3cm}{\centering \textbf{Method}}}              \vline&\textbf{\# poses}& \parbox{1.2cm}{\centering \textbf{r-1}}	& \parbox{0.7cm}{\textbf{mAP}} \Tstrut\Bstrut	\\ \hline
		\multirow{8}{*}{\parbox{1.4cm}{\centering Pose\\based}} & \multirow{4}{*}{\rotatebox{90}{Kmeans}} & 8        & 86.40	& 80.23  \Tstrut   \\
		&                         									& 12       & 88.46  & 81.59	  \\
		&                        									& 16       & 88.40  & 82.10	  \\
		&                         									& 24       & 88.10  & 82.11	\Bstrut  \\ \cline{2-5}
		                        		  & \multirow{4}{*}{\rotatebox{90}{GMM}}	& 8        & 88.21  & 81.61	 \Tstrut \\
		&                         									& 12       & \textcolor{red}{90.87}  & \textcolor{red}{82.67}	  \\
		&                         									& 16       & 90.81  & 82.43	  \\
		&                        									& 24       & 89.98  & 82.44	\Bstrut  \\ \hline
		\multirow{8}{*}{\parbox{1.4cm}{\centering Body-\\joint\\based}} & \multirow{4}{*}{\rotatebox{90}{Kmeans}} & 8        & 86.74  & 81.20	\Tstrut  \\		&                         									& 12       & 89.97  & 82.14	  \\
		&                         									& 16       & \textcolor{blue}{90.88}  & \textcolor{blue}{81.45}	  \\
		&                         									& 24       & 90.86  & 82.04	\Bstrut  \\ \cline{2-5}
		                         		  & \multirow{4}{*}{\rotatebox{90}{GMM}}	& 8        & 71.41  & 68.42	 \Tstrut \\
		&                         									& 12       & 73.49  & 70.14	  \\
		&                         									& 16       & 74.13  & 70.89	  \\
		&                         									& 24       & 74.89	& 71.43 \Bstrut	\\ \hline
	\end{tabular}
\end{table}

\subsubsection{Results on bechmark Datasets}
Our results are compared against the state-of-the-art re-ID models. The Market-1501 and DukeMTMC-reID has been widely used in the literature to validate re-ID performance, but most works only report the rank-1 accuracy and mAP score, and the same approach is being followed for the sake of comparison. However, for CUHK03 dataset, elaborate results containing rank-1, rank-5 and rank-10 accuracy alongwith mAP score is compared. Similar approach is taken while comparing our results in CUHK01 dataset where we compare our results on rank-1, rank-5 and rank-10 accuracy.

The accuracy comparison in Table ~\ref{table:market-1501}, ~\ref{table:dukemtmc}, ~\ref{table:cuhk03} and ~\ref{table:cuhk01} establish that our model is on par with the state-of-the-art methods. Our model has shown significant success not only in large datasets, but in smaller datasets (CUHK01) too. Also, it consistently outperforms all the GAN based methods developed for re-ID ~\cite{pngan}~\cite{fdgan}~\cite{dukemtmc-reid-usgan}~\cite{liu2018pose}. The significant contribution of this work is the increase over the baseline. We have achieved an improvement of $\sim 9.64\%$ (rank-1) and $\sim 16.03\%$ (mAP) in the Market-1501 dataset, which is significantly higher compared to the existing works. In view of only rank-1 accuracy, the gain over baseline is $\sim12.80\%$, $\sim10.73\%$ and $\sim13.96\%$ in DukeMTMC-reID, CUHK03 and CUHK01 dataset, respectively.

Table~\ref{table:improvement} shows the relative improvement over baseline for the state-of-the-art models that have reported higher rank-1 accuracy values than the proposed model. The recent works show a higher accuracy mostly due to the incorporation of external features such as attention maps, semantic parsing etc. Most of these works follow a different philosophy and use a higher baseline while developing the model, thereby achieving higher accuracy. However, the proposed model excels in terms of percentage increment over baseline, primarily due to two factors: 
\begin{enumerate}
	\item Data augmentation for the pt-GAN model to gain independence from rotation, occlusion, illumination and scale; and 
	\item Feature extraction by combining the multi-pose (i.e. multi-view) images, generated by the pt-GAN model.
\end{enumerate}

\begin{figure*}[htpb]
	\centering
	\includegraphics[width=0.95\linewidth]{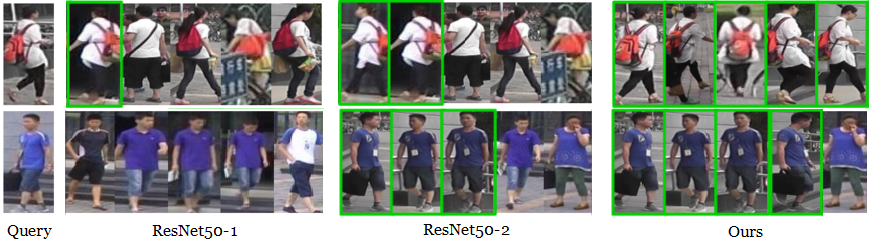}
	\caption{Comparison of retrieved images for 2 sample query image. The correct retrievals are denoted with green border. The Proposed model (ours) shows better accuracy compared to the two baseline models.}
	\label{fig:result_image_demo}
\end{figure*}

The improvement is also reflected in the retrieved images. Figure~\ref{fig:result_image_demo} shows the retrieved images for 2 sample query image for ResNet50-1, ResNet50-2 and the proposed method. Since the ResNet50-2 model has been finetuned on reid datasets, its performance is better compared to vanilla ResNet50-1. However, the Proposed model shows the best accuracy among the three.

\begin{table}[htpb]
	\centering
	\caption{Comparison of results in the Market-1501 dataset. The best and second best results are denoted with red and blue, respectively.}
	\label{table:market-1501}
	\begin{tabular}{c||c c|c c}
		\hline
		\multirow{3}{*}{\parbox{2.5cm}{\centering \textbf{Method}}} & \multicolumn{4}{c}{\parbox{3.6cm}{\centering \textbf{Market-1501}}} \Tstrut \Bstrut \\ \cline{2-5}
						& \multicolumn{2}{c}{\parbox{1.8cm}{\centering \textbf{Single Query}}} & \multicolumn{2}{c}{\parbox{1.8cm}{\centering \textbf{Multi Query}}} \Tstrut \Bstrut \\ \cline{2-5}
						& \parbox{0.9cm}{\centering \textbf{r-1}}	& \parbox{0.9cm}{\centering \textbf{mAP}} & \parbox{0.9cm}{\centering \textbf{r-1}}	& \parbox{0.8cm}{\centering \textbf{mAP}} \Tstrut \Bstrut \\ \hline
		
		BOW~\cite{market-1501}					& 44.42		& 20.76 	& - 		& -  	\\
		LSTM Siamese~\cite{varior2016siamese} 	& - 		& - 		& 61.60		& 35.31	\\
		Gated Siamese~\cite{varior2016gated}	& 65.88		& 39.55		& 76.50 	& 48.50	\\
		SpindleNet~\cite{spindlenet}			& 76.90 	& - 		& - 		& -		\\
		HP-Net~\cite{liu2017hydraplus}			& 76.90 	& - 		& - 		& -		\\
		PIE (poseBox)~\cite{posebox}			& 79.33 	& 55.95 	& - 		& -		\\
		DLPAR~\cite{zhao2017deeply} 			& 81.00 	& 63.40 	& - 		& - 	\\
		SSM~\cite{bai2017scalable} 				& 82.21 	& 68.80 	& 88.2 		& 76.2	\\
		PDC~\cite{su2017pose} 			    	& 84.14 	& 63.41 	& - 		& -		\\
		SVDNet (RE)~\cite{randomerasing}		& 87.08 	& 71.31 	& - 		& -		\\
		DML~\cite{zhang2018deep}				& 87.70 	& 68.80 	& - 		& -		\\
		DeepTransfer~\cite{geng2016deep}		& 83.70 	& 65.50 	& 89.60 	& 73.80	\\
		JLML~\cite{li2017person}				& 85.10 	& 65.50 	& 89.70 	& 73.80	\\
		MLFN~\cite{chang2018multi}				& 90.00 	& 74.30 	& - 		& -		\\
		HA-CNN~\cite{li2018harmonious} 			& 91.20 	& 75.70 	& - 		& -		\\
		PCB~\cite{weinrich2013appearance}		& \textcolor{red}{93.80} & 81.60	& - 	& -		\\
		\hline
		US-GAN~\cite{dukemtmc-reid-usgan}		& 83.97 	& 66.07 	& 88.42 	& 76.10	\\
		Liu et al.~\cite{liu2018pose}			& 87.65 	& 68.92 	& - 		& -		\\
		PN-GAN~\cite{pngan}						& 89.43 	& 72.58 	& 92.93 	& 80.19	\\
		FD-GAN~\cite{fdgan}						& 90.50 	& 77.70 	& - 		& -		\\
		\hline
		ResNet50-1 (Baseline-1)					& 83.69		& 76.48 	& -			& -		\\
		ResNet50-2 (Baseline-2) 				& 87.24 	& 80.19 	& - 		& -		\\
		\hline
		Ours 									& 90.87 	& \textcolor{blue}{82.67} 	& \textcolor{blue}{92.98} 	& \textcolor{blue}{88.32}	\\
		Ours (rerank) 							& \textcolor{blue}{91.76} 	& \textcolor{red}{88.74} 	& \textcolor{red}{93.64} 	& \textcolor{red}{89.24}	\\
		\hline
		
	\end{tabular}
\end{table}

\begin{table}[htpb]
	\centering
	\caption{Comparison of results in the DukeMTMC-reID dataset. The best and second best results are denoted with red and blue, respectively.}
	\label{table:dukemtmc}
	\begin{tabular}{c||c c}
		\hline
		\multirow{2}{*}{\parbox{3.5cm}{\centering \textbf{Method}}} & \multicolumn{2}{c}{\parbox{3.6cm}{\centering \textbf{DukeMTMC-reID}}} \Tstrut \Bstrut \\ \cline{2-3}
		& \parbox{1.3cm}{\centering \textbf{r-1}}	& \parbox{1.2cm}{\centering \textbf{mAP}} \Tstrut \Bstrut \\ \hline
		
		BoW~\cite{market-1501}			& 25.13			& 12.17	\\
		PAN~\cite{zheng2018pedestrian}	& 71.59			& 51.51	\\
		FMN~\cite{ding2017let}			& 74.51			& 56.88	\\
		SVDNet~\cite{sun2017svdnet}		& 76.7			& 56.8	\\
		HA-CNN~\cite{li2018harmonious}  & 80.5			& 63.8	\\
		Deep-person~\cite{bai2020deep}	& 80.9			& 64.8	\\
		MLFN~\cite{chang2018multi}		& 81.2			& 62.8	\\
		PCB~\cite{weinrich2013appearance}& 83.3			& 69.2	\\
		Part-aligned\cite{suh2018part}	& \textcolor{red}{84.4}		& \textcolor{blue}{69.3}	\\
		\hline
		US-GAN~\cite{dukemtmc-reid-usgan}& 67.68 		& 47.13	\\
		PN-GAN~\cite{pngan}				& 73.58 		& 53.2	\\
		Liu et al.~\cite{liu2018pose}	& 78.52 		& 56.91	\\
		FD-GAN~\cite{fdgan}				& 80.0 			& 64.5	\\
		\hline
		ResNet50-1 (Baseline-1) 		& 74.48 		& 58.95	\\
		ResNet50-2 (Baseline-2) 		& 78.04 		& 60.56	\\
		\hline
		Ours 							& 83.46 		& 68.10	\\
		Ours (rerank) 					& \textcolor{blue}{84.01} 		& \textcolor{red}{69.94}	\\
		\hline
		
	\end{tabular}
\end{table}

\begin{table}[htpb]
	\centering
	\caption{Comparison of results in the CUHK03 dataset. The best and second best results are denoted with red and blue, respectively.}
	\label{table:cuhk03}
	\begin{tabular}{c||c c c c}
		\hline
		\multirow{2}{*}{\parbox{2.5cm}{\centering \textbf{Method}}} & \multicolumn{4}{c}{\parbox{3.6cm}{\centering \textbf{CUHK03}}} \Tstrut \Bstrut \\ \cline{2-5}
		& \parbox{0.9cm}{\centering \textbf{r-1}}	& \parbox{0.9cm}{\centering \textbf{r-5}} & \parbox{0.9cm}{\centering \textbf{r-10}}	& \parbox{0.8cm}{\centering \textbf{mAP}} \Tstrut \Bstrut \\ \hline

		DeepReid~\cite{deepreid}		        & 19.89 & 50.0 	& 64.0 	& -	\\
		LSTM Siamese~\cite{varior2016siamese}   & 57.3 	& 80.1 	& 88.3 	& -	\\
		PIE~\cite{posebox}				        & 67.1 	& 92.2 	& 96.6 	& -	\\
		Gated Siamese~\cite{varior2016gated}    & 68.1 	& 88.1 	& 94.6 	& -	\\
		PDC~\cite{su2017pose}   				& 78.92 & 94.83 & 97.15 & -	\\
		DLPAR~\cite{zhao2017deeply}				& 81.6 	& 97.3 	& 98.56 & -	\\
		SpindleNet~\cite{spindlenet}			& 88.5 	& \textcolor{blue}{97.8} 	& 98.6 	& -	\\
		SVDNet~\cite{sun2017svdnet}				& 81.8 	& - 	& - 	& 84.8	\\
		JLML~\cite{li2017person}				& 83.2 	& \textcolor{red}{98.0} 	& \textcolor{red}{99.4} 	& -	\\
		\hline
		US-GAN~\cite{dukemtmc-reid-usgan}		& 84.6 	& 97.6 	& \textcolor{blue}{98.9} 	& 87.4	\\
		PN-GAN~\cite{pngan}     				& 79.76 & 93.79 & 98.56 & -	\\
		FD-GAN~\cite{fdgan}         			& \textcolor{red}{92.6} 	& - 	& - 	& \textcolor{red}{91.3}	\\
		\hline
		ResNet50-1 (Baseline-1) 		& 83.30 & 88.54 & 92.13 & 76.51	\\
		ResNet50-2 (Baseline-2) 		& 85.00 & 92.51 & 96.84 & 80.10	\\
		\hline
		Ours 							& 91.56 & 96.14 & 98.09 & 90.69 \\
		Ours (rerank) 					& \textcolor{blue}{92.24} & 97.35 & 98.86 & \textcolor{blue}{91.14}	\\
		\hline
				
	\end{tabular}
\end{table}

\begin{table}[htpb]
	\centering
	\caption{Comparison of results in the CUHK01 dataset. The best and second best results are denoted with red and blue, respectively.}
	\label{table:cuhk01}
	\begin{tabular}{c||c c c}
		\hline
		\multirow{2}{*}{\parbox{2.5cm}{\centering \textbf{Method}}} & \multicolumn{3}{c}{\parbox{3cm}{\centering \textbf{CUHK01}}} \Tstrut \Bstrut \\ \cline{2-4}
		& \parbox{0.9cm}{\centering \textbf{r-1}}	& \parbox{0.9cm}{\centering \textbf{r-5}} & \parbox{0.9cm}{\centering \textbf{r-10}} \Tstrut \Bstrut \\ \hline
		
		Ahmed et al.~\cite{ahmed2015improved}	& 47.53 & 71.5 	& 80.0 \\
		DeepRanking~\cite{chen2016deep}			& 50.41 & 75.93 & 84.07	\\
		Ensembles~\cite{paisitkriangkrai2015learning}& 53.4 	& 76.3 	& 84.4	\\
		ImpTrpLoss~\cite{cheng2016person}		& 53.7 	& 84.3 	& 91.0	\\
		GOG~\cite{matsukawa2016hierarchical}	& 57.8 	& 79.1 	& 86.2	\\
		Quadruplet~\cite{chen2017beyond}		& 62.55 & 83.44 & 89.71	\\
		NullReid~\cite{zhang2016learning}		& 69.09 & 86.87 & 91.77	\\
		PersonNet~\cite{wu2016personnet}		& 71.1 	& 90.1 	& 95.0	\\
		SpindleNet~\cite{spindlenet}			& \textcolor{red}{79.9} 	& \textcolor{red}{94.4} 	& \textcolor{red}{97.1}	\\
		\hline
		PN-GAN~\cite{pngan}						& 67.65 & 86.64 & 91.82	\\
		\hline
		ResNet50-1 (Baseline-1) 				& 64.89 & 83.76 & 89.84	\\
		ResNet50-2 (Baseline-2) 				& 67.82 & 86.07 & 91.57	\\
		\hline
		Ours 									& 73.2 	& 91.25 & 96.24	\\
		Ours (rerank) 							& \textcolor{blue}{73.95} & \textcolor{blue}{93.04} & \textcolor{blue}{96.97}	\\
		\hline

	\end{tabular}
\end{table}

\begin{table}[htpb]
	\centering
	\caption{Comparison of Rank-1 accuracy with the latest State-of-the-art models in Person ReID. The improvement (\%) over the baseline  have been taken as the metric of comparison, where the proposed model excels.}
	\label{table:improvement}
	\begin{tabular}{c||c|ccc}
		\hline
		\multicolumn{5}{c}{Rank-1} \Tstrut \Bstrut \\ 
		\hline
		\parbox{.8cm}{\centering \textbf{Dataset}}     & {\parbox{1.5cm}{\centering \textbf{Method}}} & \parbox{1cm}{\centering \textbf{Baseline}}  & \parbox{1cm}{\centering \textbf{Final}}          & \parbox{1.5cm}{\centering \textbf{Improvement}}    \Tstrut \Bstrut\\
		\hline
		\multirow{11}{*}{\rotatebox{90}{Market-1501}} 
		& Auto-ReID~\cite{autoreid}          & 94.8           & 95.4           & 0.63           \\
		& pf-SAN~\cite{pf-SAN}             & 93.8           & 94.7           & 0.96           \\
		& AdaptiveReID~\cite{adaptivereid} & 94.6           & 96             & 1.48           \\
		& VA-ReID~\cite{vareid}            & 94.7           & 96.79          & 2.21           \\
		& ABD-net~\cite{abdnet}            & 91.5           & 95.6           & 4.48           \\
		& part-aligned~\cite{suh2018part}       & 88.8           & 93.4           & 5.18           \\
		& AlignedReiD~\cite{alignedreid}        & 86.3           & 91.8           & 6.37           \\
		& DCDS~\cite{dcds}               & 87.5           & 94.1           & 7.54           \\
		& st-ReID~\cite{streid}            & 91.2           & 98.1           & 7.57           \\
		& PCB~\cite{sun2018beyond}                & 86.7           & 93.8           & 8.19           \\
		\cline{2-5}
		& \textbf{Ours}      & 83.69 & 91.76 & \textbf{9.64}  \\
		\hline
		\multirow{6}{*}{\rotatebox{90}{\begin{minipage}{1.5cm}\centering DukeMTMC-reID\end{minipage}}}                 
		& AdaptiveReID~\cite{adaptivereid} & 88.00          & 92.20          & 4.77           \\
		& pf-SAN~\cite{pf-SAN}             & 83.30          & 89.00          & 6.84           \\
		& VA-ReID~\cite{vareid}            & 87.39          & 93.85          & 7.39           \\
		& ABD-net~\cite{abdnet}            & 82.80          & 89.00          & 7.49           \\
		& st-ReID~\cite{streid}            & 83.80          & 94.40          & 12.65          \\
		\cline{2-5}
		& \textbf{Ours}      & 74.48 & 84.01 & \textbf{12.80} \\
		\hline
		\multirow{5}{*}{\rotatebox{90}{CUHK03}}       
		& AlignedReiD~\cite{alignedreid}        & 83.8           & 92.4           & 10.26          \\
		& DCDS~\cite{dcds}               & 87.7           & 95.8           & 9.24           \\
		& pf-SAN~\cite{pf-SAN}             & 63.7           & 69.7           & 9.42           \\
		& Auto-ReID~\cite{autoreid}          & 75             & 77.9           & 3.87           \\
		\cline{2-5}
		& \textbf{Ours}      & 83.3  & 92.24 & \textbf{10.73} \Tstrut \Bstrut \\ \hline 
	\end{tabular}
\end{table}		

\subsubsection{Robustness}
\begin{figure}[htpb]
	\centering
	\includegraphics[width=0.9\linewidth]{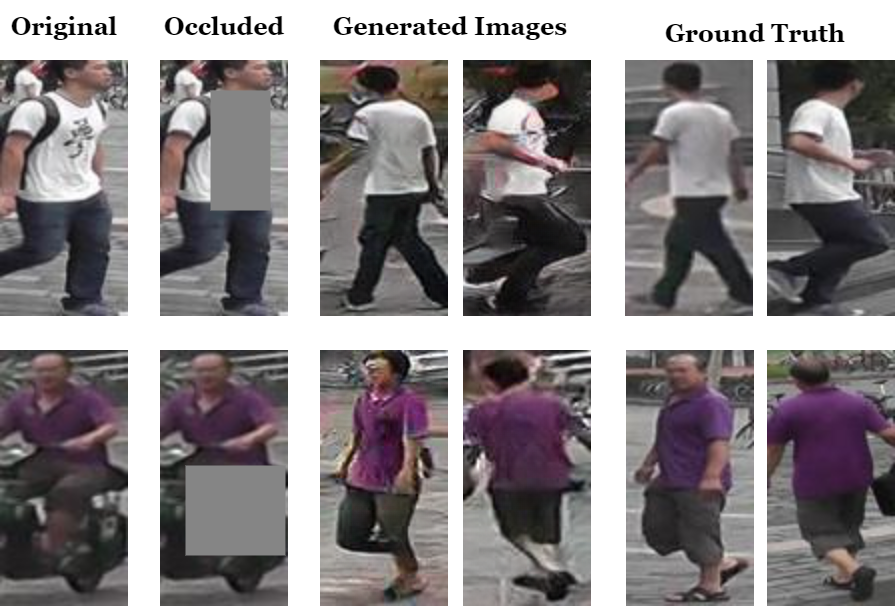}
	\caption{Occlusion invariance in the proposed pt-GAN model. The occluded image is provided as the input to the model and the generated images are compared to the ground truth. The pt-GAN model is able to successfully reconstruct the occluded regions in various poses.}
	\label{fig:occlusion}
\end{figure}

Our pt-GAN model shows good robustness towards occlusion, illumination and scale. As seen in the person 1 of Figure ~\ref{fig:gandemo}, the occlusion (railings) present in the input image is not propagated in any of the generated images. Fig.~\ref{fig:occlusion} demonstrates the occlusion invariance in the proposed model. The occluded images are given as the input to the model and are successfully reconstructed in the desired poses. Also, our model can handle human detection errors or human crop errors i.e. scaling issues. As depicted in person 2 of Figure ~\ref{fig:gandemo}, the pt-GAN model is able to correctly locate the human in the frame to extract only person-specific features for reconstruction.

\begin{figure}[htpb]
	\centering
	\includegraphics[width=0.8\linewidth]{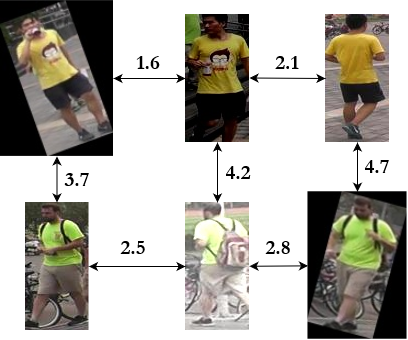}
	\caption{Comparative study of feature distances in the presence of rotation and illumination variation. The intra-class feature distances is significantly lower compared to that of the inter-class feature distance. The feature distances are scaled by a value of $10^3$}
	\label{fig:robustness}
\end{figure}

A comparative study of feature distances is carried out to understand the intra-class and inter-class variations of extracted features in Figure~\ref{fig:robustness}. The intra-class feature distances is significantly lower, even in the presence of rotation and illumination variation as well as pose change; compared to that of the inter-class feature distances. Therefore, we achieve invariance in terms of pose, occlusion, scaling, rotation and illumination with our end-to-end re-ID model.

\subsubsection{Failure Cases}
A few inconsistencies in the image generation part has been observed. Although our model generates qualitatively good images, the very fine details such as details of the face and the graphics of the t-shirt, are not properly reconstructed, which is primarily because of the low resolution of the input image.

\section{Conclusion}\label{conclusion}
This paper proposes a novel person re-identification pipeline using GANs. The image generation capability of pose transformation GAN (pt-GAN) is used to model the subject under selected poses, which is defined by the Pose Clustering module, and then features from all these representations are extracted and combined using the Feature Fusion Network. This end-to-end feature extraction model is used to extract a single viewpoint-invariant feature for each query, and the ranking is performed using feature similarity. Our model performs on par with the state-of-the-art models and outperforms the higher accuracy models in terms of improvement over baseline. We believe that our work will be useful to the community where unsupervised learning methods such as GANs can benefit person re-ID applications.

The possibility of further improving the overall re-ID performance using a different metric for matching i.e. retrieval is yet to be studied. The incorporation of attention maps or part-based information in addition to the proposed method can also be a potential future work.

%



\ifCLASSOPTIONcaptionsoff
  \newpage
\fi



%

\bibliographystyle{IEEEtran}
\bibliography{bibliography.bib}

%
%

%








\end{document}